\pdfoutput=1

\documentclass[11pt]{article}

\usepackage[]{acl}

\usepackage{times}
\usepackage{latexsym}

\usepackage[T1]{fontenc}

\usepackage[utf8]{inputenc}

\usepackage{microtype}

\usepackage{amssymb}
\usepackage{booktabs}
\usepackage{amsmath}
\usepackage{breqn}
\usepackage{graphicx}
\usepackage{enumitem}
\usepackage[explicit]{titlesec}
\usepackage{soul}
\usepackage{xcolor}
\definecolor{blue2}{RGB}{205, 255, 181}
\usepackage{tikz}
\sethlcolor{blue2}
\usetikzlibrary{shapes}
\usepackage{colortbl}
\newcommand{\mytriangle}[1]{\tikz{\node[draw=#1,fill=#1,isosceles
triangle,isosceles triangle stretches,shape border rotate=90,minimum
width=0.15cm,minimum height=0.1cm,inner sep=0pt] at (0,0) {};}}

\usepackage{tabularx}
\usepackage{pifont}
\newcommand{\xmark}{\ding{55}}%
\usepackage{xcolor}
\usepackage{subcaption} 
\usepackage{microtype}
\newcommand\cincludegraphics[2][]{\raisebox{-0.3\height}{\includegraphics[#1]{#2}}}
\usepackage[noorphans,vskip=0ex]{quoting}

\usepackage{xurl}

\title{Improving Multimodal Classification of Social Media Posts by Leveraging Image-Text Auxiliary Tasks}

\author{Danae S\'{a}nchez Villegas\textsuperscript{1} \quad Daniel Preo\c{t}iuc-Pietro\textsuperscript{2} \quad Nikolaos Aletras\textsuperscript{1} \\
        \textsuperscript{1}University of Sheffield, \textsuperscript{2}Bloomberg \\
        \texttt{\{dsanchezvillegas1, n.aletras\}@sheffield.ac.uk} \\
        \texttt{dpreotiucpie@bloomberg.net} \\
}

\begin{document}
\maketitle

\begin{abstract}
Effectively leveraging multimodal information from social media posts is essential to various downstream tasks such as sentiment analysis, sarcasm detection or hate speech classification. Jointly modeling text and images is challenging because cross-modal semantics might be hidden or the relation between image and text is weak. However, prior work on multimodal classification of social media posts has not yet addressed these challenges. 
In this work, we present an extensive study on the effectiveness of using two auxiliary losses jointly with the main task during fine-tuning multimodal models. 
First, Image-Text Contrastive (ITC) is designed to minimize the distance between image-text representations within a post, thereby effectively bridging the gap between posts where the image plays an important role in conveying the post's meaning. Second, Image-Text Matching (ITM) enhances the model's ability to understand the semantic relationship between images and text, thus improving its capacity to handle ambiguous or loosely related modalities. We combine these objectives with five multimodal models across five diverse social media datasets, demonstrating consistent improvements of up to 2.6 points F1. Our comprehensive analysis shows the specific scenarios where each auxiliary task is most effective.\footnote{Code is available here: \url{https://github.com/danaesavi/SocialMedia-TextImage-Classification-AuxLosses}.}
\end{abstract}


\section{Introduction}
Multimodal content including text and images is prevalent in social media platforms~\cite{vempala-preotiuc-pietro-2019-categorizing,sanchez-villegas-aletras-2021-point}. The content of both text and images has been widely used to improve upon single modality approaches in various downstream tasks such as sentiment analysis \cite{niu2016sentiment,ju-etal-2021-joint,tian-etal-2023-end}, hate speech and rumor detection \cite{zhao2021comparative,hossain-etal-2022-mute,cao-etal-2022-prompting,ocampo-etal-2023-playing,mu-etal-2023-time} and sarcasm detection \cite{xu-etal-2020-reasoning,liang-etal-2022-multi,ao-etal-2022-combining,tian-etal-2023-dynamic}.

\renewcommand{\arraystretch}{1.1}
\begin{figure}[!t]
    \scriptsize
    \centering
     \begin{tabular}{m{1cm}m{2cm}m{2cm}}
        \begin{tabular}[l]{@{}l@{}}   {\bf Post} \end{tabular} &
        \cincludegraphics[scale=0.12]{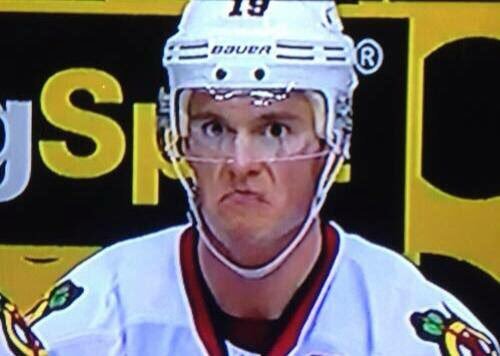}
        & \cincludegraphics[scale=0.082]{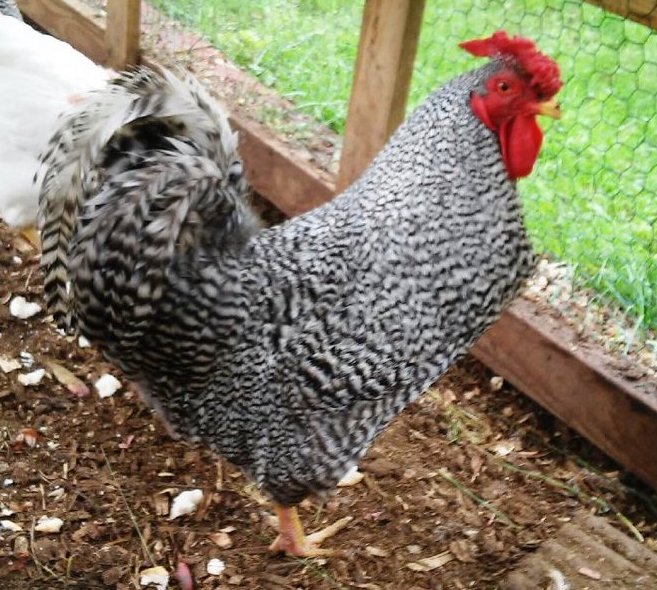} 
    \\ 
        &When @USER gets more followers than you in 12 hours
        
       &
         My baby approves
        
    \\ \hline
        \begin{tabular}[l]{@{}l@{}}{\bf Image-Text} \\ {\bf Relation}\end{tabular}
         &The image adds to the meaning 
         
         & The image does not add to the meaning 
        
    \\ \hline
        {\bf Caption}
        & A close up of a hockey player wearing a helmet
        & 
        A gray and white chicken standing in the dirt
    \end{tabular}

    \caption{Image-text relations in social media posts from \citet{vempala-preotiuc-pietro-2019-categorizing} and corresponding image captions generated with InstructBLIP. While image captions have a clear visual-language connection, image-text relationships in social media posts may no be apparent.
    }
    \label{fig:examples}
\end{figure}

Multimodal classification methods for social media tasks often combine text and image representations obtained from pre-trained models. These are usually pre-trained on standard vision-language data such as image captions where strong image-text connections are assumed, i.e., captions that explicitly describe a corresponding image \cite{hessel-lee-2020-multimodal,xu-li-2022-borrowing}. Modeling text-image pairs from social media posts presents additional challenges. A notable difficulty lies in effectively capturing latent cross-modal semantics that may not be apparent. Figure \ref{fig:examples} (left) shows an example where the text refers specifically to the mood of the person in the photo (i.e., ``unhappy feeling'' \textit{when @USER gets more followers...}). Moreover, cases where the visuals are weakly related to the text are also prevalent \cite{xu-etal-2022-understanding}. For instance, Figure \ref{fig:examples} (right) shows an image of a hen accompanied by the text \textit{My baby approves}. It is difficult to draw a direct relationship between the two without any additional context.

Multimodal models for social media classification can be divided into: (1) \textit{single-stream} models where image and text features are concatenated together and fed into the same module such as Unicoder \cite{li2020unicoder}, VisualBERT \cite{li2019visualbert}, ViLT \cite{kim2021vilt} and ALPRO \cite{li2022align}; and (2) \textit{dual-stream} approaches where images and text are processed separately, e.g., ViLBert \cite{lu2019vilbert}, LXMERT \cite{tan-bansal-2019-lxmert}, METER \cite{dou2022empirical} and BLIP-2 \cite{li2023blip}.
Consequently, these models might still suffer from the aforementioned issues. 

In this work, we examine the use of two tasks -- Image-Text Contrastive (ITC) and Image-Text Matching (ITM) -- as auxiliary losses during fine-tuning for improving social media post classification. By using the ITC contrastive loss \cite{he2020momentum,li2021align,yu2022coca}, we anticipate that when the image contributes to the post's meaning, as illustrated in Fig. \ref{fig:examples} (left), the model will place them closer in the representation space. Conversely, ITM leverages binary classification loss for image-text alignment \cite{chen2019uniter,tan-bansal-2019-lxmert,wang2021ufo}. We expect that this will improve the model's ability to handle posts where associations may not be explicitly stated as shown in Fig. \ref{fig:examples} (right). Although ITC and ITM have been used as pre-training objectives using generic images and their corresponding captions \cite{radford2021learning,wang2021ufo,chen2022altclip}, their potential for enhancing fine-tuning in social media classification has yet to be explored. 

Our main contributions are as follows: (1) we present an extensive study on comparing multimodal models jointly fine-tuned with ITC and ITM covering both \textit{single-} and \textit{dual-stream} approaches; (2) we show that models using ITC and ITM as auxiliary losses consistently improve their performance across five diverse multimodal social media  datasets; (3) we offer a comprehensive analysis revealing the effectiveness of individual auxiliary tasks and their combination across various image-text relationship types in posts.

\begin{table*}[t!]
\centering
\scriptsize
\begin{tabularx}{0.95\textwidth}{llcXXXX}
\toprule
\textbf{Dataset} & \textbf{Classification Task} & \textbf{\#} &\textbf{Train} & \textbf{Val} & \textbf{Test} & \textbf{All}  \\ \midrule

TIR \cite{vempala-preotiuc-pietro-2019-categorizing}   & \begin{tabular}[c]{@{}l@{}} Text-Image Relation\\ Classification \end{tabular} & 4  & 3,575          & 447          & 449           & 4,471      \\ \hline

MVSA \cite{niu2016sentiment}  & Sentiment Analysis & 3 & 3,611  & 451  & 451  & 4,511  \\ \hline

MHP  \cite{gomez2020exploring,botelho-etal-2021-deciphering} & \begin{tabular}[c]{@{}l@{}}Hate Speech \\ Classification \end{tabular} & 4 & 3,998          & 500          & 502           & 5,000     \\  \hline

MSD  \cite{cai-etal-2019-multi}   & Sarcasm Detection & 2 & 19,816         & 2,410        & 2,409         & 24,635   \\  \hline

MICD  \cite{villegas2023multimodal}   & \begin{tabular}[c]{@{}l@{}} Influencer Commercial \\ Content Detection \end{tabular}& 2 &  11,377        & 1,572         & 1,435 & 14,384  \\ 

\bottomrule
\end{tabularx}
\caption{Description and statistics of each dataset. \# refers to number of classes.}
\label{tab:datasets}
\end{table*}
\section{Multimodal Auxiliary Tasks}
\label{sec:aux-tasks}

\paragraph{Image-Text Contrastive (ITC)} 

Modeling text-image pairs in social media posts involves capturing hidden cross-modal semantics \citep{vempala-preotiuc-pietro-2019-categorizing,kruk-etal-2019-integrating}. For instance, in Figure \ref{fig:examples} (left) the visible mood of the person on the photo is related to the text of the post. Instead of directly matching images with textual descriptions (e.g., \textit{a man wearing a helmet}), we aim to encourage the model to capture the dependencies between the image and text within the posts. 

For this purpose, we use the ITC objective \cite{he2020momentum,li2021align,yu2022coca} which pushes towards a feature space in which image and text representations of a post are brought closer together, while image and text representations that appear in different posts are pushed further apart. 
Let $L_n$ and $I_n$ be the n-th (normalized) representation of text and accompanying image of a post in a training batch. While the cosine similarity of the pair $L_n$ and $I_n$ is minimized, the cosine similarity of all other random pairs (e.g., $L_n$ and $I_m$; $I_m$ is an image from a different post in the current batch) is maximized. Given $N$ posts within a training batch, ITC loss is defined as follows: 

\scriptsize 
\begin{gather} 
l_{ITC} = \frac{1}{2} (l_1 + l_2) \\
l_1 = -\frac{1}{N} \Sigma_{n=1}^N log \frac{exp(LI^T/e^\tau)}{\Sigma_{j=1}^N exp(LI^T/e^\tau)} \\
l_2 = -\frac{1}{N} \Sigma_{n=1}^N log \frac{exp(IL^T/e^\tau)}{\Sigma_{j=1}^N exp(IL^T/e^\tau)} 
\end{gather}
\normalsize
\noindent $\tau$ is a learnable temperature parameter to scale the logits \citep{jia2021scaling}.

\paragraph{Image-Text Matching (ITM)}

In social media posts, unrelated or weakly related text-image pairs are common \cite{hessel-lee-2020-multimodal, xu-etal-2022-understanding} such as the post depicted in Fig. \ref{fig:examples} (right). 
To address this, we use the ITM objective \cite{chen2019uniter,tan-bansal-2019-lxmert,wang2021ufo} during fine-tuning to understand the semantic correspondence between images and text. ITM involves a binary classification loss that penalizes the model when a given text and image do not appear together in a post. Let $I_n$ and $L_n$ be the image and text representation of the n-th post in a training batch, we randomly replace $I_n$ with an image of another post from the current batch with a probability of $0.5$ following \cite{wang2021ufo,kim2021vilt}. If $I_n$ is replaced, then the image and text do not match, otherwise $I_n$ and $L_n$ match. Thus, the ITM loss corresponds to the cross-entropy loss for penalizing incorrect predictions, $l_{ITM} = -\Sigma_{i=1}^2 t_i log(p_i)$ 
where $t_i$ is the gold label (matched or mismatched) and $p_i$ is the softmax probability for each label. 
 
\paragraph{Joint Fine-tuning Objectives}

The joint fine-tuning loss function includes the cross-entropy classification loss  ($l_{CE}$) and the two auxiliary training objectives 
defined as: $l_{C+M} = \lambda_1 l_{CE} + \lambda_2 l_{ITC} + \lambda_3 l_{ITM}$, where $\lambda_1,\lambda_2,\lambda_3$ are hyperparameters to control the influence of each loss. 

\section{Experimental Setup}
\label{sec:exp-setup}

\subsection{Datasets}
\label{sec:datasets}
We experiment with five diverse multimodal public datasets in English: (1) \textbf{TIR} -- text-image relationship categorization \cite{vempala-preotiuc-pietro-2019-categorizing}; (2) \textbf{MVSA} -- multi-view sentiment analysis \cite{niu2016sentiment}; (3) \textbf{MHP} -- multimodal hate speech detection \cite{gomez2020exploring,botelho-etal-2021-deciphering}; (4) \textbf{MSD} -- multimodal sarcasm detection \cite{cai-etal-2019-multi}: and (5) \textbf{MICD} -- multimodal commercial influencer content detection \citep{villegas2023multimodal}. Table \ref{tab:datasets} presents dataset statistics. 

\subsection{Single Modality Methods}

\paragraph{Text-only} We fine-tune \textbf{BERT} \cite{devlin-etal-2019-bert} and \textbf{Bernice} \citep{delucia-etal-2022-bernice}, a BERT based model pre-trained on a corpus of multilingual tweets. We also experiment with few-shot (FS) prompting using \textbf{Flan-T5} \citep{chung2022scaling} and \textbf{GPT-3} \cite{brown2020language}. For each dataset, we construct a few-shot prompt and include two randomly selected training examples for each class.\footnote{Appx. \ref{appdx:prompting} shows the prompt templates.}

\paragraph{Image-only} We fine-tune \textbf{ResNet}152 \cite{he2016deep} and \textbf{ViT} \cite{dosovitskiy2020image}, both pre-trained on ImageNet \cite{russakovsky2015imagenet}. We experiment with few-shot prompting using \textbf{IDEFICS} \citep{laurenccon2023obelisc} and zero-shot prompting using \textbf{InstructBLIP} \citep{dai2023instructblip}. Prompts include two randomly chosen image-only training examples per class (see Appx. \ref{appdx:prompting}).

\subsection{Multimodal Models}
\label{sec:exp-multimodal-models}

\paragraph{Ber-ViT} 
We use Bernice and ViT to obtain representations of the text ($L$) and image ($I$). 
\textbf{Ber-ViT-Conc} appends the text and image vectors from the corresponding $L$ and $I$ \small{[CLS]} ~\normalsize tokens to obtain the multimodal representation $h^{LI}$; 
\textbf{Ber-ViT-Att} computes cross-attention between $L$ and $I$. $h^{LI}$ is obtained by appending the \small{[CLS]} ~\normalsize token from $L$ and the \small{[CLS]} ~\normalsize token from the attention layer. We fine-tune each model by adding a classification layer. 

\paragraph{MMBT} \cite{kiela2019supervised}. Image embeddings obtained from Resnet152 are concatenated with token embeddings and passed to a BERT-like transformer. The \small{[CLS]} ~\normalsize token is used as the multimodal representation ($h^{LI}$) for classification. 

\paragraph{LXMERT} \citep{tan-bansal-2019-lxmert} consists of three encoders and their corresponding outputs for vision $I$, language $L$, and a multimodal vector $h^{LI}$. 

\paragraph{ViLT} We fine-tune ViLT \citep{dosovitskiy2020image} and extract the multimodal  $h^{LI}$ that corresponds to the first token from the last hidden state.

\paragraph{ITC and ITM Inputs} The ITC auxiliary task inputs are the corresponding text and image vectors of each model. The ITM auxiliary task input is the respective multimodal representation $h^{LI}$.

\subsection{Evaluation} Results are obtained over three runs using different random seeds reporting average and standard deviation. We use weighted F1 for model evaluation following standard practice on the TIR, MHP and MICD datasets to manage class imbalance.\footnote{Implementation details are included in Appx. \ref{sec:implem-details}.}

\section{Results}

\subsection{Performance Comparison} 

\paragraph{Image-text auxiliary tasks improve multimodal classification.}
Table \ref{tab:results-mm} shows that multimodal models surpass single-modality approaches across all datasets. We consistently find performance gains when using either ITC, ITM, or both auxiliary losses during fine-tuning, with improvements up to 2.6 F1 over each base model. Therefore, we can improve performance without costly pre-training on social media text-image tasks. These findings are especially valuable in multimodal computational social science studies, where grasping the interplay between text and images is vital \cite{sanchez-villegas-etal-2021-analyzing,xu-etal-2022-understanding}.

\renewcommand{\arraystretch}{1.2}
\begin{table}[t!]
\scriptsize
\centering
\begin{tabularx}{\columnwidth}{|l|X|X|X|X|X|c|}
\hline
\textbf{Model} & \textbf{TIR}   & \textbf{MVSA}  & \textbf{MHP} & \textbf{MSD}  & \textbf{MICD} & \mytriangle{black}\\ \hline
\multicolumn{1}{|l|}{Majority Class}       & 16.0            & 59.8         & 53.4        & 45.2 &48.0& -\\ \hline \hline

\multicolumn{7}{|l|}{\textbf{Text-only Models}}        \\ \hline
\multicolumn{1}{|l|}{BERT}              & 37.2$_{1.3}$        & 70.1$_{0.8}$         & 73.3$_{1.3}$        & 83.9$_{0.2}$ &74.3$_{0.6}$&     -  \\ 
\multicolumn{1}{|l|}{Bernice}              & 38.9$_{1.1}$       & 71.6$_{0.6}$         & 73.6$_{0.6}$        & 84.5$_{0.8}$ &74.5$_{2.2}$&    -   \\ 
\multicolumn{1}{|l|}{Flan-T5$^*$}            &    3.8$_{0.0}$        &   58.9$_{0.0}$          &      46.5$_{1.3}$       &     59.6$_{2.2}$  &48.7$_{1.6}$&  -  \\ 
\multicolumn{1}{|l|}{GPT-3$^*$}                &     16.3$_{6.1}$       &      55.9$_{0.1}$       &      58.2$_{4.6}$       & 69.6$_{2.7}$ &69.6$_{1.5}$&  - \\ 
\hline \hline

\multicolumn{7}{|l|}{\textbf{Image-only Models}}                     \\ \hline
\multicolumn{1}{|l|}{ResNet152}                  & 48.2$_{0.0}$        & 63.8$_{0.1}$       & 51.8$_{5.8}$       &        46.9$_{0.1}$  &59.6$_{0.5}$&   - \\ 
\multicolumn{1}{|l|}{ViT}                  & 51.4$_{1.3}$      & 68.2$_{0.6}$         & 57.2$_{1.2}$        &        71.5$_{0.1}$   &60.8$_{1.3}$&  - \\ 
\multicolumn{1}{|l|}{IDEFICS$^*$}                  & 12.4$_{3.6}$      & 34.7$_{6.1}$         & 34.9$_{2.7}$        &        58.9$_{2.4}$   &35.6$_{0.0}$&  - \\

\multicolumn{1}{|l|}{InstructBLIP$^*$}                  & 3.9$_{0.0}$      & 47.2$_{0.0}$         & 11.0$_{0.0}$        &        22.7$_{0.0}$   &35.6$_{0.0}$&  - 
\\ \hline \hline

\multicolumn{7}{|l|}{\textbf{Multimodal Models}}     \\ \hline

\multicolumn{1}{|l|}{Ber-ViT-Conc}             & 43.6$_{1.2}$        & 70.4$_{0.0}$         & 76.6$_{0.6}$   & 88.8$_{0.0}$  &75.5$_{1.9}$& - \\ 
\multicolumn{1}{|l|}{ +C}   & 44.9$_{0.7}$        & \hl{72.0$^\dagger_{0.2}$}         & 77.3$_{1.1}$   & \textbf{\underline{\hl{89.7}}}\hl{$^\dagger_{0.0}$} & \underline{77.2}$_{0.4}$& 1.2\\ 
\multicolumn{1}{|l|}{ +M}   & 44.1$_{0.2}$        & \underline{\hl{73.6}}\hl{$^\dagger_{0.9}$}         & \underline{77.8}$_{0.6}$   & \hl{89.2$^\dagger_{0.1}$} &76.1$_{0.8}$& 1.2\\ 
\multicolumn{1}{|l|}{ +C+M}    & \underline{45.8}$_{0.8}$        & \hl{73.4$^\dagger_{0.4}$}   &     \hl{77.7$^\dagger_{0.6}$}    & \textbf{\underline{\hl{89.7}}}\hl{$^\dagger_{0.2}$} &76.3$_{0.5}$& 1.6 \\ 
\hline

\multicolumn{1}{|l|}{Ber-ViT-Att}                   & 53.7$_{1.0}$     & 72.1$_{0.7}$    & 76.8$_{0.5}$          & 88.8$_{0.3}$ &75.6$_{0.8}$& -\\ 
\multicolumn{1}{|l|}{ +C}         & 54.8$_{0.8}$     & 72.8$_{0.2}$    & 77.5$_{0.6}$ & 89.5$^\dagger_{0.5}$ &\underline{\textbf{\hl{77.8}}}\hl{$^\dagger_{0.5}$}& 0.8\\ 
\multicolumn{1}{|l|}{ +M}         & \textbf{\underline{\hl{55.9}}}\hl{$^\dagger_{0.8}$}     & \hl{73.5$^\dagger_{0.2}$}    & 77.4$_{0.6}$ &  89.4$_{0.5}$ &76.6$_{0.5}$&  1.2\\ 
\multicolumn{1}{|l|}{+C+M}   & 54.6$_{0.7}$     & \underline{\textbf{\hl{74.6}}}\hl{$^\dagger_{0.3}$}  & \underline{\textbf{\hl{78.0}}}\hl{$^\dagger_{0.1}$} &  \underline{\textbf{\hl{89.7}}}\hl{$^\dagger_{0.3}$} &
76.3$_{0.2}$& 1.7\\ 
\hline

\multicolumn{1}{|l|}{MMBT}                  & 53.2$_{1.2}$    & 72.4$_{0.4}$   & 74.5$_{0.5}$    & 83.2$_{0.0}$ &73.6$_{0.4}$& -\\ 
\multicolumn{1}{|l|}{ +C}         & \underline{53.7}$_{1.1}$    & 73.2$_{1.0}$   & 75.7$_{1.7}$    & \underline{\hl{84.4}}\hl{$^\dagger_{0.3}$}  &
74.1$_{0.8}$& 1.1\\ 
\multicolumn{1}{|l|}{ +M}         & \underline{53.7}$_{0.7}$    & 73.4$_{0.8}$   & 75.4$_{1.3}$    & \hl{84.3$^\dagger_{0.3}$} & \underline{\hl{74.8}}\hl{$^\dagger_{0.6}$}&  0.9\\ 
\multicolumn{1}{|l|}{ +C+M}   & 53.6$_{0.2}$    & \underline{\hl{73.5}}\hl{$^\dagger_{0.0}$}   & \underline{75.7}$_{1.2}$  &  83.4$_{0.2}$  & 73.8$_{0.5}$&  0.6 \\ 
\hline

\multicolumn{1}{|l|}{LXMERT}           & 51.3$_{0.5}$          & 68.2$_{1.1}$          & 70.7$_{0.8}$       &  81.9$_{0.5}$ &69.9$_{1.0}$& - \\ 
\multicolumn{1}{|l|}{ +C}    & 51.9$_{0.3}$         & \underline{\hl{70.4}}\hl{$^\dagger_{0.5}$}   & \underline{\hl{72.1}}\hl{$^\dagger_{0.2}$}      &  \underline{82.7}$_{0.1}$ &70.8$_{0.5}$& 1.2\\ 
\multicolumn{1}{|l|}{ +M}    & 51.8$_{0.4}$             & 69.5$_{0.2}$      & 71.8$_{0.8}$      &   82.3$_{0.5}$ &\underline{70.9}$_{0.2}$& 0.9 \\ 
\multicolumn{1}{|l|}{ +C+M}    & \underline{52.3}$_{1.4}$ & 69.3$_{0.9}$    & 71.9$_{1.7}$       &  82.1$_{0.4}$   &70.3$_{0.3}$& 0.8 \\ 
\hline

\multicolumn{1}{|l|}{ViLT}              & 53.1$_{1.1}$         & 70.5$_{1.3}$ & 71.8$_{0.0}$    & 83.0$_{0.8}$ &67.8$_{1.6}$& - \\ 
\multicolumn{1}{|l|}{ +C}    & \underline{\hl{55.7}}\hl{$^\dagger_{0.2}$} & \underline{72.9}$_{1.0}$ & \hl{72.5$^\dagger_{0.4}$}    & 83.4$_{0.4}$ &68.3$_{0.2}$& 1.3 \\ 
\multicolumn{1}{|l|}{ +M}    & \underline{\hl{55.7}}\hl{$^\dagger_{0.3}$} & 72.1$_{2.3}$ & 72.0$_{0.5}$    & \underline{83.5}$_{0.2}$ &68.7$_{1.1}$& 1.1\\ 
\multicolumn{1}{|l|}{ +C+M}    & \hl{55.3$^\dagger_{0.3}$}  & \underline{72.9}$_{1.3}$ & \underline{73.4}$_{1.4}$    & 83.2$_{0.4}$  &
\underline{70.0}$_{1.3}$& 1.7 \\
\hline
\end{tabularx}
\caption[]{Results in weighted F1 for all datasets. 
Best results for each base multimodal model are underlined and best results for each dataset are in bold. \hl{$^\dagger$} indicates statistically significant improvement (t-test, $p < 0.05$) over the corresponding base model. Subscripts denote standard deviation over three runs. \mytriangle{black} refers to the average relative improvement over each base model across datasets.$^*$ denotes prompting. 
+C,+M, C+M refer to +ITC, +ITM and +ITC+ITM.
}
\label{tab:results-mm}
\end{table}

\paragraph{Dual-stream methods are effective in leveraging information from the auxiliary tasks.} Across MVSA, MHP and MSD datasets, the Ber-ViT-Att+C+M model achieves the best performance ($74.6$, $78.0$, and $89.7$ F1 respectively). Generally, we observe that both ITC and ITM contribute to the performance improvements of Ber-ViT-Att. Overall, Ber-ViT-Att+C and Ber-ViT-Att+M models average improvements over the base model across datasets are 0.8 and 1.2 respectively, while Ber-ViT-Att+C+Mimprovement is 1.7. 
The performance gap between \textit{dual-} and \textit{single-stream} models is narrower in TIR. ViLT+M achieves $55.7$ F1 while Ber-ViT-Att+M obtains $55.9$. This is likely due to the importance of visual information for this task (i.e., predicting the semiotic relationship between images and text), which is better aligned with ViLT as a visual-based model.

\subsection{Training with different number of samples}

\begin{figure*}
    \centering
    \includegraphics[width=0.95\linewidth]{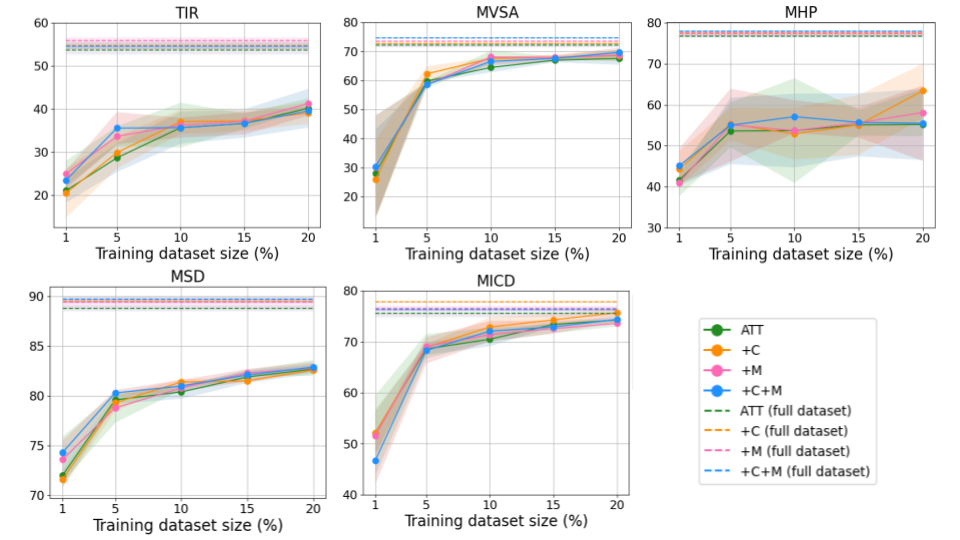}
    \caption{Results in weighted F1 using Ber-ViT-Att (ATT) for all datasets when training with different percentages of training data. We plot the mean and standard deviation across three runs. }
    \label{fig:low-res}
\end{figure*}

To test the generalizability and data efficiency of our models, we conduct experiments using our best performing model, Ber-ViT-Att, across different training data sizes, thus simulating low resource scenarios. We assessed the weighted F1 scores of Ber-ViT-Att both independently and with the incorporation of each auxiliary loss, as well as a combination of both. The results of these experiments are presented in Figure \ref{fig:low-res}. 
While Table \ref{tab:results-mm}, highlights that the highest performance is generally achieved using both auxiliary losses, in Figure \ref{fig:low-res} we observe the best performing models are predominantly distributed between Ber-ViT-Att+C and Ber-ViT-Att+C+M.

We find that the difference between training with $20\%$ of random examples and using the entire dataset is modest in some cases, particularly when fine-tuning with both ITC and ITM losses on MVSA, MSD, and MICD. Specifically, for MSD the difference is $6.8$ F1 points, while for MVSA and MICD, it is less than $5$ F1 points. These results suggest that our models exhibit robust generalization. However, MHP exhibits a more substantial difference, with a gap of $21.6$ F1 points when Ber-ViT-Att is trained with $20\%$ of the training examples, narrowing to $14.1$ F1 points with Ber-ViT-Att+C. This suggests the viability of employing ITC as an auxiliary loss during fine-tuning for hate speech classification in low-resource scenarios.

\begin{figure}[!t]
\tiny
\centering
\begin{tabular}{m{3.4cm}|m{3.4cm}}
\hline
\scriptsize{\textbf{Text is represented in image}} & \scriptsize{\textbf{Text is not represented in image}} \\ 

\textbf{Image adds to the meaning}  
& \textbf{Image adds to the meaning} 
\\
\includegraphics[scale=0.104]{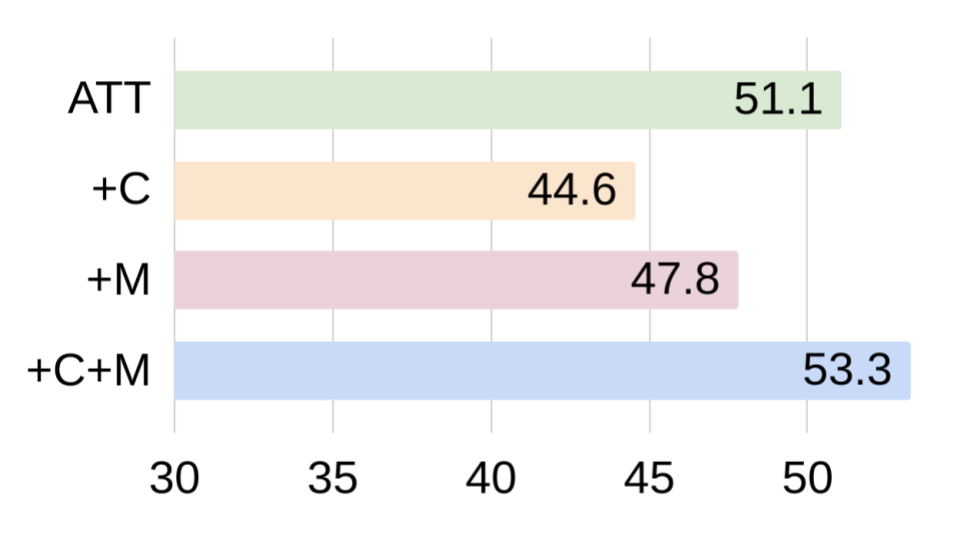}  & \includegraphics[scale=0.104]{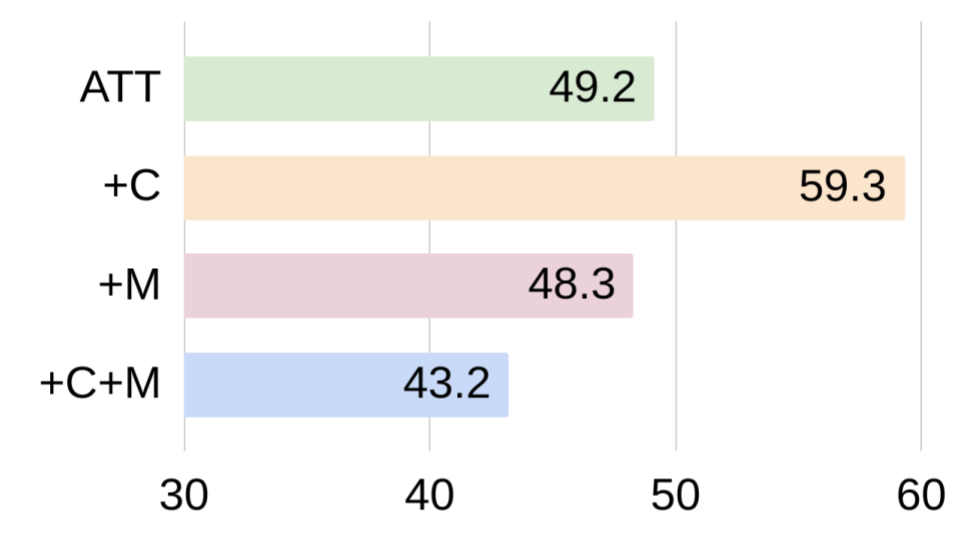}  \\ 
\hline 
\scriptsize{\textbf{Text is represented in image}} & \scriptsize{\textbf{Text is not represented in image}} \\
\textbf{Image does not add to the meaning}  
& \textbf{Image does not add to the meaning} 
\\ 

&\\
\includegraphics[scale=0.104]{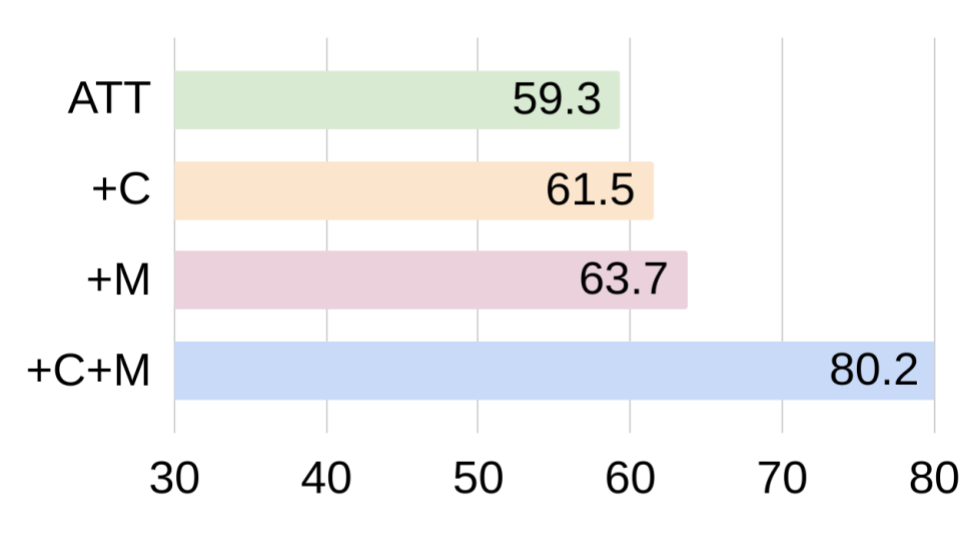}  & \includegraphics[scale=0.104]{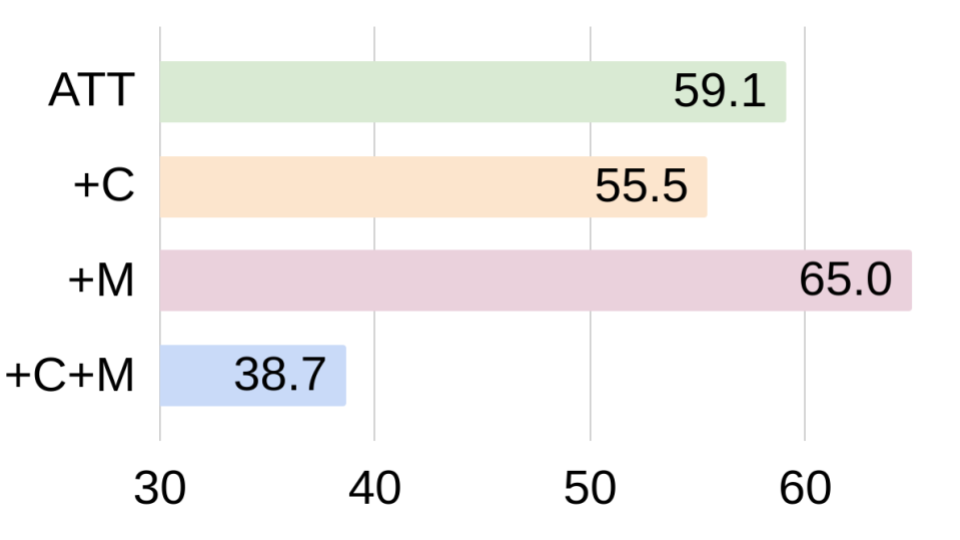} 
\\ 
\end{tabular}
\caption{Accuracy per label using Ber-ViT-Att (ATT) across different image-text relation types based on image contribution to the post's meaning and text representation on the image. 
}
\label{tab:acc-plots}
\end{figure}
\section{Analysis}

\begin{figure}[!t]
\tiny
\centering
\begin{tabular}{m{3.35cm}|m{3.35cm}}
\hline
\scriptsize{\textbf{Text is represented in image}} & \scriptsize{\textbf{Text is not represented in image}} \\  

\textbf{Image adds to the meaning}  
& \textbf{Image adds to the meaning} 
\\ 
 \includegraphics[scale=0.065]{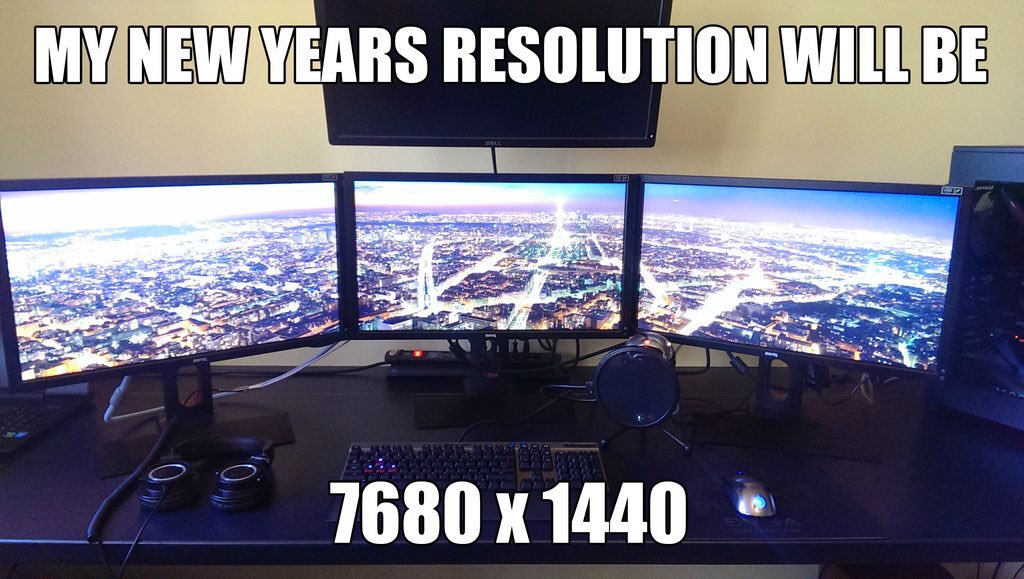} &\includegraphics[scale=0.105]{images/T701818489040818176.jpeg}    
 \\

\textit{New Years Resolution.}
& \textit{When @USER gets more followers than you in 12 hours}
\\ 

\scriptsize{ATT:\xmark{} | +C:\checkmark | +M:\checkmark | +C+M:\checkmark} 
& \scriptsize{ATT:\xmark{} | +C:\checkmark | +M:\xmark{} | +C+M:\xmark} \\
\\ \hline
\scriptsize{\textbf{Text is represented in image}} & \scriptsize{\textbf{Text is not represented in image}} \\
\textbf{Image does not add to the meaning}  
& \textbf{Image does not add to the meaning} 
\\ 
\includegraphics[scale=0.04]{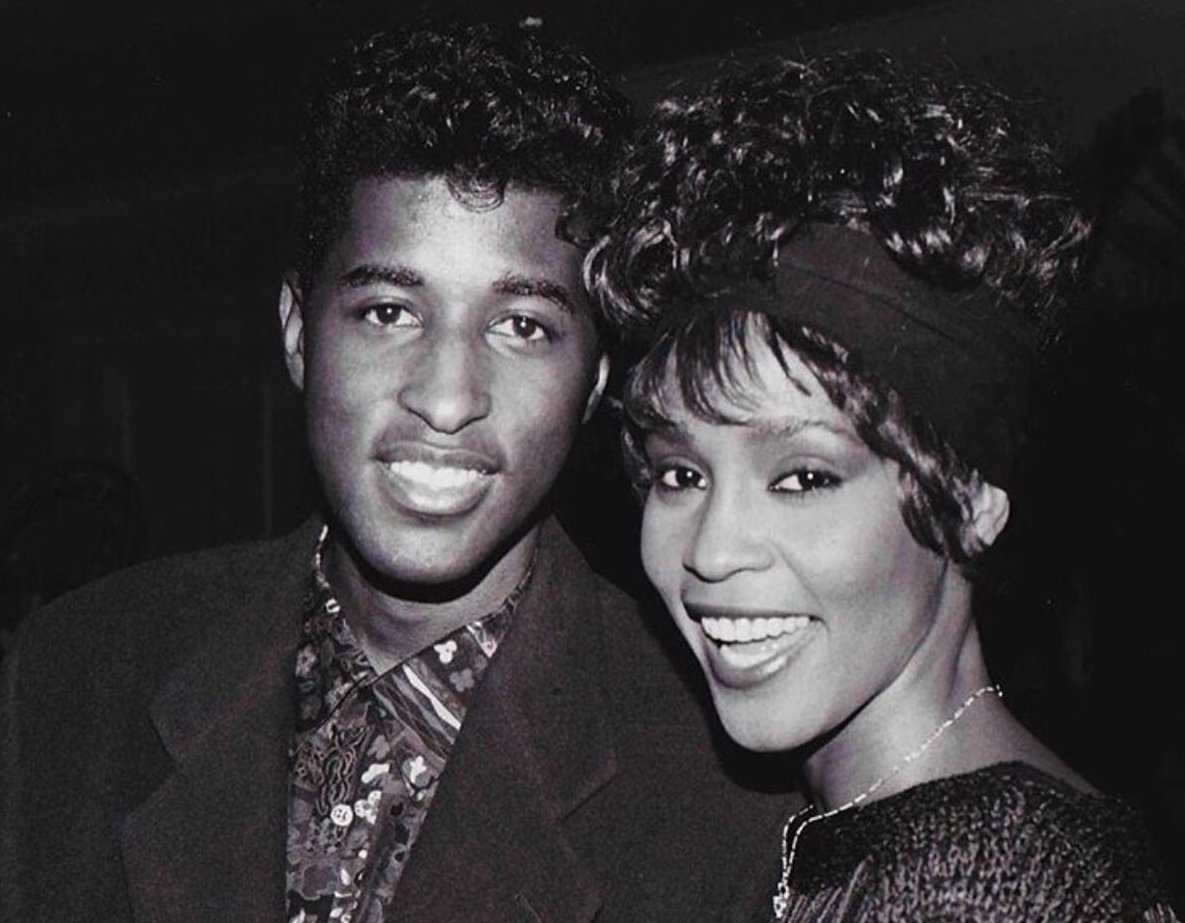}  & \includegraphics[scale=0.065]{images/T693566822222860289.jpeg} \\
\textit{Babyface and Whitney Houston}
& \textit{My baby approves}  \\ 
\\
\scriptsize{ATT:\xmark{} | +C:\xmark{} | +M:\xmark{} | +C+M:\checkmark }
& \scriptsize{ATT:\xmark{} | +C:\xmark{} | +M:\checkmark | +C+M:\xmark } \\

\end{tabular}
\caption{Bert-ViT-Att (ATT) predictions on randomly selected examples with varying image-text relations. 
}

\label{tab:examples-acc}
\end{figure}

 We analyze Ber-ViT-Att's predictions on TIR to understand when each auxiliary task benefits different image-text relations as categorized by \citet{vempala-preotiuc-pietro-2019-categorizing} based on image contribution and text representation (Figure \ref{tab:acc-plots} and \ref{tab:examples-acc}).

\noindent\textbf{When the text is represented in the image} using both auxiliary tasks (models denoted with \small{+C+M}),~\normalsize the model achieves the best performance, especially when the visual content is not semantically relevant to the post. We observe that $80.2$\% of the tweets are correctly classified achieving a substantial improvement over the Ber-ViT-Att baseline where only $59.3$\% of the posts are correctly classified. 
 
 \noindent\textbf{When text is not represented on the image,} we find that including \small{ITC}~\normalsize performs best when the visual content is relevant, with $59.3$\% of the tweets correctly classified compared to $49.2$\% using Ber-ViT-Att. Finally, in cases where the image does not enhance the semantic meaning, Ber-ViT-Att+M exhibits the highest performance, correctly classifying $65$\% of the posts. This validates our hypothesis that incorporating ITM helps models to effectively identify posts with weaker image-text relationships. 

\section{Conclusion}
We presented an extensive study on the effectiveness of using two auxiliary tasks, Image-Text Contrastive and Image-Text Matching when fine-tuning multimodal models for social media posts classification. This approach addresses the challenges of hidden cross-modal semantics and weak image-text relationships in social media content. 
 

\section*{Limitations}
First, the datasets used in our experiments are solely in English. This choice allows for consistency and comparability across the datasets, but it does not test the generalizability of our findings to other languages. In future work, we plan to extend our research to a multilingual setting to address this limitation. 
The effectiveness of the models incorporating auxiliary tasks depends on the underlying base model, however, our approach can easily be adapted to new models. Finally, the inclusion of auxiliary tasks in our models introduces an increase in training time. For instance, the training time for Ber-ViT-Att on the TIR dataset is approximately 1.5 hours on an Nvidia A100 GPU. When incorporating the auxiliary tasks (Ber-ViT-Att+C+M), the training time extends to around 2.5 hours, a 66\% relative increase in training time. However, the additional time is a one-time occurrence and relatively minor when compared to the pre-training times of large language models (LLMs).

\textbf{Experiments on TIR dataset}. We align with previous work on the TIR dataset by employing text-only and image-only models for classification \citep{vempala-preotiuc-pietro-2019-categorizing}, with the expectation that specific textual cues or image content can indicate relationships, even without considering the image content. For instance, (a) tweets concluding with an ellipsis or brief comments may serve as predictive indicators that the text is not represented in the accompanying image, and (b) images featuring people may be more likely to contain text corresponding to the names of those individuals. While unimodal models may not be ideal choices in real-world scenarios for this task, they serve as valuable performance baseline. 


\section*{Acknowledgments}
DSV and NA are supported by the Leverhulme Trust under Grant Number: RPG\#2020\#148. DSV is also supported by the Centre for Doctoral Training in Speech and Language Technologies (SLT) and their Applications funded by the UK Research and Innovation grant EP/S023062/1. We would like to thank Katerina Margatina, Mali Jin, Constantinos Karouzos, and all reviewers for their valuable feedback.
\bibliography{anthology,custom}

\begin{thebibliography}{57}
\expandafter\ifx\csname natexlab\endcsname\relax\def\natexlab#1{#1}\fi

\bibitem[{Anderson et~al.(2018)Anderson, He, Buehler, Teney, Johnson, Gould, and Zhang}]{anderson2018bottom}
Peter Anderson, Xiaodong He, Chris Buehler, Damien Teney, Mark Johnson, Stephen Gould, and Lei Zhang. 2018.
\newblock Bottom-up and top-down attention for image captioning and visual question answering.
\newblock In \emph{Proceedings of the IEEE conference on computer vision and pattern recognition}, pages 6077--6086.

\bibitem[{Ao et~al.(2022)Ao, Sanchez~Villegas, Preotiuc-Pietro, and Aletras}]{ao-etal-2022-combining}
Xiao Ao, Danae Sanchez~Villegas, Daniel Preotiuc-Pietro, and Nikolaos Aletras. 2022.
\newblock \href {https://doi.org/10.18653/v1/2022.naacl-main.131} {Combining humor and sarcasm for improving political parody detection}.
\newblock In \emph{Proceedings of the 2022 Conference of the North American Chapter of the Association for Computational Linguistics: Human Language Technologies}, pages 1800--1807, Seattle, United States. Association for Computational Linguistics.

\bibitem[{Bird and Loper(2004)}]{bird-loper-2004-nltk}
Steven Bird and Edward Loper. 2004.
\newblock \href {https://aclanthology.org/P04-3031} {{NLTK}: The natural language toolkit}.
\newblock In \emph{Proceedings of the {ACL} Interactive Poster and Demonstration Sessions}, pages 214--217, Barcelona, Spain. Association for Computational Linguistics.

\bibitem[{Botelho et~al.(2021)Botelho, Hale, and Vidgen}]{botelho-etal-2021-deciphering}
Austin Botelho, Scott Hale, and Bertie Vidgen. 2021.
\newblock \href {https://doi.org/10.18653/v1/2021.findings-acl.166} {Deciphering implicit hate: Evaluating automated detection algorithms for multimodal hate}.
\newblock In \emph{Findings of the Association for Computational Linguistics: ACL-IJCNLP 2021}, pages 1896--1907, Online. Association for Computational Linguistics.

\bibitem[{Brown et~al.(2020)Brown, Mann, Ryder, Subbiah, Kaplan, Dhariwal, Neelakantan, Shyam, Sastry, Askell et~al.}]{brown2020language}
Tom Brown, Benjamin Mann, Nick Ryder, Melanie Subbiah, Jared~D Kaplan, Prafulla Dhariwal, Arvind Neelakantan, Pranav Shyam, Girish Sastry, Amanda Askell, et~al. 2020.
\newblock Language models are few-shot learners.
\newblock \emph{Advances in neural information processing systems}, 33:1877--1901.

\bibitem[{Cai et~al.(2019)Cai, Cai, and Wan}]{cai-etal-2019-multi}
Yitao Cai, Huiyu Cai, and Xiaojun Wan. 2019.
\newblock \href {https://doi.org/10.18653/v1/P19-1239} {Multi-modal sarcasm detection in {T}witter with hierarchical fusion model}.
\newblock In \emph{Proceedings of the 57th Annual Meeting of the Association for Computational Linguistics}, pages 2506--2515, Florence, Italy. Association for Computational Linguistics.

\bibitem[{Cao et~al.(2022)Cao, Lee, Chong, and Jiang}]{cao-etal-2022-prompting}
Rui Cao, Roy Ka-Wei Lee, Wen-Haw Chong, and Jing Jiang. 2022.
\newblock \href {https://doi.org/10.18653/v1/2022.emnlp-main.22} {Prompting for multimodal hateful meme classification}.
\newblock In \emph{Proceedings of the 2022 Conference on Empirical Methods in Natural Language Processing}, pages 321--332, Abu Dhabi, United Arab Emirates. Association for Computational Linguistics.

\bibitem[{Chen et~al.(2020)Chen, Li, Yu, El~Kholy, Ahmed, Gan, Cheng, and Liu}]{chen2019uniter}
Yen-Chun Chen, Linjie Li, Licheng Yu, Ahmed El~Kholy, Faisal Ahmed, Zhe Gan, Yu~Cheng, and Jingjing Liu. 2020.
\newblock Uniter: Learning universal image-text representations.
\newblock \emph{European Conference on Computer Vision}.

\bibitem[{Chen et~al.(2022)Chen, Liu, Zhang, Ye, Yang, and Wu}]{chen2022altclip}
Zhongzhi Chen, Guang Liu, Bo-Wen Zhang, Fulong Ye, Qinghong Yang, and Ledell Wu. 2022.
\newblock Altclip: Altering the language encoder in clip for extended language capabilities.
\newblock \emph{arXiv preprint arXiv:2211.06679}.

\bibitem[{Chung et~al.(2022)Chung, Hou, Longpre, Zoph, Tay, Fedus, Li, Wang, Dehghani, Brahma et~al.}]{chung2022scaling}
Hyung~Won Chung, Le~Hou, Shayne Longpre, Barret Zoph, Yi~Tay, William Fedus, Eric Li, Xuezhi Wang, Mostafa Dehghani, Siddhartha Brahma, et~al. 2022.
\newblock Scaling instruction-finetuned language models.
\newblock \emph{arXiv preprint arXiv:2210.11416}.

\bibitem[{Dai et~al.(2023)Dai, Li, Li, Tiong, Zhao, Wang, Li, Fung, and Hoi}]{dai2023instructblip}
Wenliang Dai, Junnan Li, Dongxu Li, Anthony Meng~Huat Tiong, Junqi Zhao, Weisheng Wang, Boyang Li, Pascale Fung, and Steven Hoi. 2023.
\newblock \href {http://arxiv.org/abs/2305.06500} {Instructblip: Towards general-purpose vision-language models with instruction tuning}.

\bibitem[{DeLucia et~al.(2022)DeLucia, Wu, Mueller, Aguirre, Resnik, and Dredze}]{delucia-etal-2022-bernice}
Alexandra DeLucia, Shijie Wu, Aaron Mueller, Carlos Aguirre, Philip Resnik, and Mark Dredze. 2022.
\newblock \href {https://doi.org/10.18653/v1/2022.emnlp-main.415} {Bernice: A multilingual pre-trained encoder for {T}witter}.
\newblock In \emph{Proceedings of the 2022 Conference on Empirical Methods in Natural Language Processing}, pages 6191--6205, Abu Dhabi, United Arab Emirates. Association for Computational Linguistics.

\bibitem[{Devlin et~al.(2019)Devlin, Chang, Lee, and Toutanova}]{devlin-etal-2019-bert}
Jacob Devlin, Ming-Wei Chang, Kenton Lee, and Kristina Toutanova. 2019.
\newblock \href {https://doi.org/10.18653/v1/N19-1423} {{BERT}: Pre-training of deep bidirectional transformers for language understanding}.
\newblock In \emph{Proceedings of the 2019 Conference of the North {A}merican Chapter of the Association for Computational Linguistics: Human Language Technologies, Volume 1 (Long and Short Papers)}, pages 4171--4186, Minneapolis, Minnesota. Association for Computational Linguistics.

\bibitem[{Dosovitskiy et~al.(2020)Dosovitskiy, Beyer, Kolesnikov, Weissenborn, Zhai, Unterthiner, Dehghani, Minderer, Heigold, Gelly et~al.}]{dosovitskiy2020image}
Alexey Dosovitskiy, Lucas Beyer, Alexander Kolesnikov, Dirk Weissenborn, Xiaohua Zhai, Thomas Unterthiner, Mostafa Dehghani, Matthias Minderer, Georg Heigold, Sylvain Gelly, et~al. 2020.
\newblock An image is worth 16x16 words: Transformers for image recognition at scale.
\newblock \emph{arXiv preprint arXiv:2010.11929}.

\bibitem[{Dou et~al.(2022)Dou, Xu, Gan, Wang, Wang, Wang, Zhu, Zhang, Yuan, Peng et~al.}]{dou2022empirical}
Zi-Yi Dou, Yichong Xu, Zhe Gan, Jianfeng Wang, Shuohang Wang, Lijuan Wang, Chenguang Zhu, Pengchuan Zhang, Lu~Yuan, Nanyun Peng, et~al. 2022.
\newblock An empirical study of training end-to-end vision-and-language transformers.
\newblock In \emph{Proceedings of the IEEE/CVF Conference on Computer Vision and Pattern Recognition}, pages 18166--18176.

\bibitem[{Gomez et~al.(2020)Gomez, Gibert, Gomez, and Karatzas}]{gomez2020exploring}
Raul Gomez, Jaume Gibert, Lluis Gomez, and Dimosthenis Karatzas. 2020.
\newblock Exploring hate speech detection in multimodal publications.
\newblock In \emph{Proceedings of the IEEE/CVF winter conference on applications of computer vision}, pages 1470--1478.

\bibitem[{He et~al.(2020)He, Fan, Wu, Xie, and Girshick}]{he2020momentum}
Kaiming He, Haoqi Fan, Yuxin Wu, Saining Xie, and Ross Girshick. 2020.
\newblock Momentum contrast for unsupervised visual representation learning.
\newblock In \emph{Proceedings of the IEEE/CVF conference on computer vision and pattern recognition}, pages 9729--9738.

\bibitem[{He et~al.(2016)He, Zhang, Ren, and Sun}]{he2016deep}
Kaiming He, Xiangyu Zhang, Shaoqing Ren, and Jian Sun. 2016.
\newblock Deep residual learning for image recognition.
\newblock In \emph{Proceedings of the IEEE conference on computer vision and pattern recognition}, pages 770--778.

\bibitem[{Hessel and Lee(2020)}]{hessel-lee-2020-multimodal}
Jack Hessel and Lillian Lee. 2020.
\newblock \href {https://doi.org/10.18653/v1/2020.emnlp-main.62} {Does my multimodal model learn cross-modal interactions? it{'}s harder to tell than you might think!}
\newblock In \emph{Proceedings of the 2020 Conference on Empirical Methods in Natural Language Processing (EMNLP)}, pages 861--877, Online. Association for Computational Linguistics.

\bibitem[{Hossain et~al.(2022)Hossain, Sharif, and Hoque}]{hossain-etal-2022-mute}
Eftekhar Hossain, Omar Sharif, and Mohammed~Moshiul Hoque. 2022.
\newblock \href {https://aclanthology.org/2022.aacl-srw.5} {{MUTE}: A multimodal dataset for detecting hateful memes}.
\newblock In \emph{Proceedings of the 2nd Conference of the Asia-Pacific Chapter of the Association for Computational Linguistics and the 12th International Joint Conference on Natural Language Processing: Student Research Workshop}, pages 32--39, Online. Association for Computational Linguistics.

\bibitem[{Jia et~al.(2021)Jia, Yang, Xia, Chen, Parekh, Pham, Le, Sung, Li, and Duerig}]{jia2021scaling}
Chao Jia, Yinfei Yang, Ye~Xia, Yi-Ting Chen, Zarana Parekh, Hieu Pham, Quoc Le, Yun-Hsuan Sung, Zhen Li, and Tom Duerig. 2021.
\newblock Scaling up visual and vision-language representation learning with noisy text supervision.
\newblock In \emph{International Conference on Machine Learning}, pages 4904--4916. PMLR.

\bibitem[{Ju et~al.(2021)Ju, Zhang, Xiao, Li, Li, Zhang, and Zhou}]{ju-etal-2021-joint}
Xincheng Ju, Dong Zhang, Rong Xiao, Junhui Li, Shoushan Li, Min Zhang, and Guodong Zhou. 2021.
\newblock \href {https://doi.org/10.18653/v1/2021.emnlp-main.360} {Joint multi-modal aspect-sentiment analysis with auxiliary cross-modal relation detection}.
\newblock In \emph{Proceedings of the 2021 Conference on Empirical Methods in Natural Language Processing}, pages 4395--4405, Online and Punta Cana, Dominican Republic. Association for Computational Linguistics.

\bibitem[{Kiela et~al.(2019)Kiela, Bhooshan, Firooz, Perez, and Testuggine}]{kiela2019supervised}
Douwe Kiela, Suvrat Bhooshan, Hamed Firooz, Ethan Perez, and Davide Testuggine. 2019.
\newblock Supervised multimodal bitransformers for classifying images and text.
\newblock \emph{arXiv preprint arXiv:1909.02950}.

\bibitem[{Kim et~al.(2021)Kim, Son, and Kim}]{kim2021vilt}
Wonjae Kim, Bokyung Son, and Ildoo Kim. 2021.
\newblock Vilt: Vision-and-language transformer without convolution or region supervision.
\newblock In \emph{International Conference on Machine Learning}, pages 5583--5594. PMLR.

\bibitem[{King and Zeng(2001)}]{king2001logistic}
Gary King and Langche Zeng. 2001.
\newblock Logistic regression in rare events data.
\newblock \emph{Political analysis}, 9(2):137--163.

\bibitem[{Kingma and Ba(2014)}]{kingma2014adam}
Diederik~P Kingma and Jimmy Ba. 2014.
\newblock Adam: A method for stochastic optimization.
\newblock \emph{arXiv preprint arXiv:1412.6980}.

\bibitem[{Kruk et~al.(2019)Kruk, Lubin, Sikka, Lin, Jurafsky, and Divakaran}]{kruk-etal-2019-integrating}
Julia Kruk, Jonah Lubin, Karan Sikka, Xiao Lin, Dan Jurafsky, and Ajay Divakaran. 2019.
\newblock \href {https://doi.org/10.18653/v1/D19-1469} {Integrating text and image: Determining multimodal document intent in {I}nstagram posts}.
\newblock In \emph{Proceedings of the 2019 Conference on Empirical Methods in Natural Language Processing and the 9th International Joint Conference on Natural Language Processing (EMNLP-IJCNLP)}, pages 4622--4632, Hong Kong, China. Association for Computational Linguistics.

\bibitem[{Lauren{\c{c}}on et~al.(2023)Lauren{\c{c}}on, Saulnier, Tronchon, Bekman, Singh, Lozhkov, Wang, Karamcheti, Rush, Kiela et~al.}]{laurenccon2023obelisc}
Hugo Lauren{\c{c}}on, Lucile Saulnier, L{\'e}o Tronchon, Stas Bekman, Amanpreet Singh, Anton Lozhkov, Thomas Wang, Siddharth Karamcheti, Alexander~M Rush, Douwe Kiela, et~al. 2023.
\newblock Obelisc: An open web-scale filtered dataset of interleaved image-text documents.
\newblock \emph{arXiv preprint arXiv:2306.16527}.

\bibitem[{Li et~al.(2022)Li, Li, Li, Niebles, and Hoi}]{li2022align}
Dongxu Li, Junnan Li, Hongdong Li, Juan~Carlos Niebles, and Steven~CH Hoi. 2022.
\newblock Align and prompt: Video-and-language pre-training with entity prompts.
\newblock In \emph{Proceedings of the IEEE/CVF Conference on Computer Vision and Pattern Recognition}, pages 4953--4963.

\bibitem[{Li et~al.(2020)Li, Duan, Fang, Gong, and Jiang}]{li2020unicoder}
Gen Li, Nan Duan, Yuejian Fang, Ming Gong, and Daxin Jiang. 2020.
\newblock Unicoder-vl: A universal encoder for vision and language by cross-modal pre-training.
\newblock In \emph{Proceedings of the AAAI Conference on Artificial Intelligence}, volume~34, pages 11336--11344.

\bibitem[{Li et~al.(2023)Li, Li, Savarese, and Hoi}]{li2023blip}
Junnan Li, Dongxu Li, Silvio Savarese, and Steven Hoi. 2023.
\newblock Blip-2: Bootstrapping language-image pre-training with frozen image encoders and large language models.
\newblock \emph{arXiv preprint arXiv:2301.12597}.

\bibitem[{Li et~al.(2021)Li, Selvaraju, Gotmare, Joty, Xiong, and Hoi}]{li2021align}
Junnan Li, Ramprasaath Selvaraju, Akhilesh Gotmare, Shafiq Joty, Caiming Xiong, and Steven Chu~Hong Hoi. 2021.
\newblock Align before fuse: Vision and language representation learning with momentum distillation.
\newblock \emph{Advances in neural information processing systems}, 34:9694--9705.

\bibitem[{Li et~al.(2019)Li, Yatskar, Yin, Hsieh, and Chang}]{li2019visualbert}
Liunian~Harold Li, Mark Yatskar, Da~Yin, Cho-Jui Hsieh, and Kai-Wei Chang. 2019.
\newblock Visualbert: A simple and performant baseline for vision and language.
\newblock \emph{arXiv preprint arXiv:1908.03557}.

\bibitem[{Liang et~al.(2022)Liang, Lou, Li, Yang, Gui, He, Pei, and Xu}]{liang-etal-2022-multi}
Bin Liang, Chenwei Lou, Xiang Li, Min Yang, Lin Gui, Yulan He, Wenjie Pei, and Ruifeng Xu. 2022.
\newblock \href {https://doi.org/10.18653/v1/2022.acl-long.124} {Multi-modal sarcasm detection via cross-modal graph convolutional network}.
\newblock In \emph{Proceedings of the 60th Annual Meeting of the Association for Computational Linguistics (Volume 1: Long Papers)}, pages 1767--1777, Dublin, Ireland. Association for Computational Linguistics.

\bibitem[{Lu et~al.(2019)Lu, Batra, Parikh, and Lee}]{lu2019vilbert}
Jiasen Lu, Dhruv Batra, Devi Parikh, and Stefan Lee. 2019.
\newblock Vilbert: Pretraining task-agnostic visiolinguistic representations for vision-and-language tasks.
\newblock \emph{Advances in neural information processing systems}, 32.

\bibitem[{Mu et~al.(2023)Mu, Bontcheva, and Aletras}]{mu-etal-2023-time}
Yida Mu, Kalina Bontcheva, and Nikolaos Aletras. 2023.
\newblock \href {https://aclanthology.org/2023.findings-eacl.55} {It{'}s about time: Rethinking evaluation on rumor detection benchmarks using chronological splits}.
\newblock In \emph{Findings of the Association for Computational Linguistics: EACL 2023}, pages 736--743, Dubrovnik, Croatia. Association for Computational Linguistics.

\bibitem[{Nguyen et~al.(2020)Nguyen, Vu, and Tuan~Nguyen}]{nguyen-etal-2020-bertweet}
Dat~Quoc Nguyen, Thanh Vu, and Anh Tuan~Nguyen. 2020.
\newblock \href {https://doi.org/10.18653/v1/2020.emnlp-demos.2} {{BERT}weet: A pre-trained language model for {E}nglish tweets}.
\newblock In \emph{Proceedings of the 2020 Conference on Empirical Methods in Natural Language Processing: System Demonstrations}, pages 9--14, Online. Association for Computational Linguistics.

\bibitem[{Niu et~al.(2016)Niu, Zhu, Pang, and El~Saddik}]{niu2016sentiment}
Teng Niu, Shiai Zhu, Lei Pang, and Abdulmotaleb El~Saddik. 2016.
\newblock Sentiment analysis on multi-view social data.
\newblock In \emph{MultiMedia Modeling: 22nd International Conference, MMM 2016, Miami, FL, USA, January 4-6, 2016, Proceedings, Part II 22}, pages 15--27. Springer.

\bibitem[{Ocampo et~al.(2023)Ocampo, Cabrio, and Villata}]{ocampo-etal-2023-playing}
Nicolas Ocampo, Elena Cabrio, and Serena Villata. 2023.
\newblock \href {https://doi.org/10.18653/v1/2023.findings-acl.173} {Playing the part of the sharp bully: Generating adversarial examples for implicit hate speech detection}.
\newblock In \emph{Findings of the Association for Computational Linguistics: ACL 2023}, pages 2758--2772, Toronto, Canada. Association for Computational Linguistics.

\bibitem[{Radford et~al.(2021)Radford, Kim, Hallacy, Ramesh, Goh, Agarwal, Sastry, Askell, Mishkin, Clark et~al.}]{radford2021learning}
Alec Radford, Jong~Wook Kim, Chris Hallacy, Aditya Ramesh, Gabriel Goh, Sandhini Agarwal, Girish Sastry, Amanda Askell, Pamela Mishkin, Jack Clark, et~al. 2021.
\newblock Learning transferable visual models from natural language supervision.
\newblock In \emph{International conference on machine learning}, pages 8748--8763. PMLR.

\bibitem[{Ren et~al.(2016)Ren, He, Girshick, and Sun}]{ren2016faster}
Shaoqing Ren, Kaiming He, Ross Girshick, and Jian Sun. 2016.
\newblock Faster {R}-{C}{N}{N}: towards real-time object detection with region proposal networks.
\newblock \emph{IEEE transactions on pattern analysis and machine intelligence}, 39(6):1137--1149.

\bibitem[{Russakovsky et~al.(2015)Russakovsky, Deng, Su, Krause, Satheesh, Ma, Huang, Karpathy, Khosla, Bernstein et~al.}]{russakovsky2015imagenet}
Olga Russakovsky, Jia Deng, Hao Su, Jonathan Krause, Sanjeev Satheesh, Sean Ma, Zhiheng Huang, Andrej Karpathy, Aditya Khosla, Michael Bernstein, et~al. 2015.
\newblock Imagenet large scale visual recognition challenge.
\newblock \emph{International journal of computer vision}, 115:211--252.

\bibitem[{S{\'a}nchez~Villegas and Aletras(2021)}]{sanchez-villegas-aletras-2021-point}
Danae S{\'a}nchez~Villegas and Nikolaos Aletras. 2021.
\newblock \href {https://doi.org/10.18653/v1/2021.emnlp-main.614} {Point-of-interest type prediction using text and images}.
\newblock In \emph{Proceedings of the 2021 Conference on Empirical Methods in Natural Language Processing}, pages 7785--7797, Online and Punta Cana, Dominican Republic. Association for Computational Linguistics.

\bibitem[{S{\'a}nchez~Villegas et~al.(2023)S{\'a}nchez~Villegas, Goanta, and Aletras}]{villegas2023multimodal}
Danae S{\'a}nchez~Villegas, Catalina Goanta, and Nikolaos Aletras. 2023.
\newblock A multimodal analysis of influencer content on twitter.
\newblock \emph{arXiv preprint arXiv:2309.03064}.

\bibitem[{S{\'a}nchez~Villegas et~al.(2021)S{\'a}nchez~Villegas, Mokaram, and Aletras}]{sanchez-villegas-etal-2021-analyzing}
Danae S{\'a}nchez~Villegas, Saeid Mokaram, and Nikolaos Aletras. 2021.
\newblock \href {https://doi.org/10.18653/v1/2021.findings-acl.321} {Analyzing online political advertisements}.
\newblock In \emph{Findings of the Association for Computational Linguistics: ACL-IJCNLP 2021}, pages 3669--3680, Online. Association for Computational Linguistics.

\bibitem[{Tan and Bansal(2019)}]{tan-bansal-2019-lxmert}
Hao Tan and Mohit Bansal. 2019.
\newblock \href {https://doi.org/10.18653/v1/D19-1514} {{LXMERT}: Learning cross-modality encoder representations from transformers}.
\newblock In \emph{Proceedings of the 2019 Conference on Empirical Methods in Natural Language Processing and the 9th International Joint Conference on Natural Language Processing (EMNLP-IJCNLP)}, pages 5100--5111, Hong Kong, China. Association for Computational Linguistics.

\bibitem[{Tian et~al.(2023{\natexlab{a}})Tian, Xu, Zhang, and Mao}]{tian-etal-2023-dynamic}
Yuan Tian, Nan Xu, Ruike Zhang, and Wenji Mao. 2023{\natexlab{a}}.
\newblock \href {https://doi.org/10.18653/v1/2023.acl-long.139} {Dynamic routing transformer network for multimodal sarcasm detection}.
\newblock In \emph{Proceedings of the 61st Annual Meeting of the Association for Computational Linguistics (Volume 1: Long Papers)}, pages 2468--2480, Toronto, Canada. Association for Computational Linguistics.

\bibitem[{Tian et~al.(2023{\natexlab{b}})Tian, Chen, Hu, Song, and Xia}]{tian-etal-2023-end}
Yuanhe Tian, Weidong Chen, Bo~Hu, Yan Song, and Fei Xia. 2023{\natexlab{b}}.
\newblock \href {https://doi.org/10.18653/v1/2023.findings-acl.859} {End-to-end aspect-based sentiment analysis with {C}ombinatory {C}ategorial {G}rammar}.
\newblock In \emph{Findings of the Association for Computational Linguistics: ACL 2023}, pages 13597--13609, Toronto, Canada. Association for Computational Linguistics.

\bibitem[{Vaswani et~al.(2017)Vaswani, Shazeer, Parmar, Uszkoreit, Jones, Gomez, Kaiser, and Polosukhin}]{Vaswani2017}
Ashish Vaswani, Noam Shazeer, Niki Parmar, Jakob Uszkoreit, Llion Jones, Aidan~N Gomez, {\L}ukasz Kaiser, and Illia Polosukhin. 2017.
\newblock Attention is all you need.
\newblock In \emph{Advances in Neural Information Processing Systems}, pages 5998--6008.

\bibitem[{Vempala and Preo{\c{t}}iuc-Pietro(2019)}]{vempala-preotiuc-pietro-2019-categorizing}
Alakananda Vempala and Daniel Preo{\c{t}}iuc-Pietro. 2019.
\newblock \href {https://doi.org/10.18653/v1/P19-1272} {Categorizing and inferring the relationship between the text and image of {T}witter posts}.
\newblock In \emph{Proceedings of the 57th Annual Meeting of the Association for Computational Linguistics}, pages 2830--2840, Florence, Italy. Association for Computational Linguistics.

\bibitem[{Wang et~al.(2021)Wang, Hu, Gan, Yang, Dai, Liu, Lu, and Wang}]{wang2021ufo}
Jianfeng Wang, Xiaowei Hu, Zhe Gan, Zhengyuan Yang, Xiyang Dai, Zicheng Liu, Yumao Lu, and Lijuan Wang. 2021.
\newblock Ufo: A unified transformer for vision-language representation learning.
\newblock \emph{arXiv preprint arXiv:2111.10023}.

\bibitem[{Wolf et~al.(2019)Wolf, Debut, Sanh, Chaumond, Delangue, Moi, Cistac, Rault, Louf, Funtowicz et~al.}]{wolf2019huggingface}
Thomas Wolf, Lysandre Debut, Victor Sanh, Julien Chaumond, Clement Delangue, Anthony Moi, Pierric Cistac, Tim Rault, R{\'e}mi Louf, Morgan Funtowicz, et~al. 2019.
\newblock Huggingface's transformers: State-of-the-art natural language processing.
\newblock \emph{arXiv preprint arXiv:1910.03771}.

\bibitem[{Xu and Li(2022)}]{xu-li-2022-borrowing}
Chunpu Xu and Jing Li. 2022.
\newblock \href {https://doi.org/10.18653/v1/2022.emnlp-main.381} {Borrowing human senses: Comment-aware self-training for social media multimodal classification}.
\newblock In \emph{Proceedings of the 2022 Conference on Empirical Methods in Natural Language Processing}, pages 5644--5656, Abu Dhabi, United Arab Emirates. Association for Computational Linguistics.

\bibitem[{Xu et~al.(2022)Xu, Tan, Li, and Li}]{xu-etal-2022-understanding}
Chunpu Xu, Hanzhuo Tan, Jing Li, and Piji Li. 2022.
\newblock \href {https://doi.org/10.18653/v1/2022.findings-emnlp.182} {Understanding social media cross-modality discourse in linguistic space}.
\newblock In \emph{Findings of the Association for Computational Linguistics: EMNLP 2022}, pages 2459--2471, Abu Dhabi, United Arab Emirates. Association for Computational Linguistics.

\bibitem[{Xu et~al.(2020)Xu, Zeng, and Mao}]{xu-etal-2020-reasoning}
Nan Xu, Zhixiong Zeng, and Wenji Mao. 2020.
\newblock \href {https://doi.org/10.18653/v1/2020.acl-main.349} {Reasoning with multimodal sarcastic tweets via modeling cross-modality contrast and semantic association}.
\newblock In \emph{Proceedings of the 58th Annual Meeting of the Association for Computational Linguistics}, pages 3777--3786, Online. Association for Computational Linguistics.

\bibitem[{Yu et~al.(2022)Yu, Wang, Vasudevan, Yeung, Seyedhosseini, and Wu}]{yu2022coca}
Jiahui Yu, Zirui Wang, Vijay Vasudevan, Legg Yeung, Mojtaba Seyedhosseini, and Yonghui Wu. 2022.
\newblock Coca: Contrastive captioners are image-text foundation models.
\newblock \emph{arXiv preprint arXiv:2205.01917}.

\bibitem[{Zhao et~al.(2021)Zhao, Zhang, and Hopfgartner}]{zhao2021comparative}
Zhixue Zhao, Ziqi Zhang, and Frank Hopfgartner. 2021.
\newblock A comparative study of using pre-trained language models for toxic comment classification.
\newblock In \emph{Companion Proceedings of the Web Conference 2021}, pages 500--507.

\end{thebibliography}

\appendix

\newpage

\section{Implementation details}
\label{sec:implem-details}

\subsection{Data Processing}
\label{sec:txtproc}
\paragraph{Text} For each tweet, we lowercase and tokenize text using the NLTK Twitter tokenizer \cite{bird-loper-2004-nltk}. We also replace URLs and user @-mentions with placeholder tokens. Emojis are replaced with their corresponding text string, e.g thumbs\_up following \citet{nguyen-etal-2020-bertweet}. 

\paragraph{Image} Images are resized to ($224\times224$) pixels representing a value for the red, green and blue color in $[0,255]$. The pixel values are normalized to $[0-1]$. For LXMERT \citep{tan-bansal-2019-lxmert} in Section \ref{sec:exp-multimodal-models}, we extract \textit{object-level} features using Faster-RCNN \cite{ren2016faster} as in \citet{anderson2018bottom} and keep $36$ objects for each image as in \citet{tan-bansal-2019-lxmert}. 

\subsection{Data Splits}
We use the same data splits for MVSA, MHP, MSD, and MICD as in the original papers. For TIR, instead of a 10-fold cross-validation, we randomly split the data in $80$\%, $10$\%, and $10$\% for training, validation, and testing for consistency with the other tasks.

\subsection{Hyperparameters}
We select the hyperparameters for all models using early stopping by monitoring the validation loss. We use the Adam optimizer~\citep{kingma2014adam}. We estimate the class weights using the `balanced' heuristic \cite{king2001logistic}. All experiments are performed using an Nvidia A100 GPU with a batch size of 8 for TIR and MHP and 16 for MVSA and MSD datasets. For prompting implementation details see Appx. \ref{appdx:prompting}.


\paragraph{Image-only} For ResNet152 \citep{he2016deep}, we fine-tune for 1, 5, 8, 6 and 1 epochs for TIR, MVSA, MHP, MSD and MICD datasets respectively, with learning rate $\eta=1e^{-5}$ and dropout $\delta=0.05$ before passing the image representation through the classification layer. We fine-tune ViT \citep{dosovitskiy2020image} for 3 epochs for TIR, MSD and MICD and 10 epochs for MVSA and MHP datasets with learning rate $\eta=1e^{-5}$ and dropout $\delta=0.05$. $\eta \in \{1e^{-3},1e^{-4},1e^{-5}\}$ and $\delta$ in $[0,0.5]$, random search.

\paragraph{Text-only Transformers} We fine-tune BERT and Bernice for 20 epochs and choose the epoch with the lowest validation loss. We use the pre-trained base-uncased model for BERT \cite{Vaswani2017, devlin-etal-2019-bert} from the Hugging Face library (12-layer, 768-dimensional) \citep{wolf2019huggingface}, and the base model for Bernice \citep{delucia-etal-2022-bernice} with a maximal sequence length of 128. We fine-tune BERT for 3, 9, 5, 2 and 1 epochs for TIR, MVSA, MHP, MSD and MICD with learning rate $\eta=1e^{-5}$ and dropout $\delta=0.05$; and Bernice for 3, 4, 7, 3 and 3 epochs for TIR, MVSA, MHP, MSD and MICD datasets, $\eta=1e^{-5}$ and $\delta=0.05$. For all models $\eta \in \{2e^{-5},1e^{-4},1e^{-5}\}$ and $\delta \in [0,0.5]$, random search.

\paragraph{Multimodal Predictive Models}
We train MMBT~\citep{kiela2019supervised}, ViLT~\citep{kim2021vilt},  LXMERT~\citep{tan-bansal-2019-lxmert} and Bernice-ViT models with $\lambda_1$, $\lambda_2$, $\lambda_3$; $\lambda_2$ and $\lambda_3 \in [0,1.5]$ (as explained in Section \ref{sec:aux-tasks}), and number of fine-tuning epochs (E) for each model as shown in Table \ref{tab:hp-lambda}. For ViLT models we keep the vision layers frozen and we use a learning rate of $\eta=1e^{-4}$, dropout $\delta=0.05$ and weight decay of 0.0002. For all other multimodal models we use a learning rate of $\eta=1e^{-5}$, dropout $\delta=0.05$ and weight decay of 0.00025. 
\begin{table*}[t!]
    \centering
    \small
    \begin{tabularx}{0.85\textwidth}{|l|X|X|l|l|}
        \hline
         \textbf{Dataset} & \textbf{Text} & \textbf{Image} & \textbf{Label} &\textbf{Outputs}\\ \hline
         
         \textbf{MVSA}
         & \scriptsize{So proud of these kids! Not only talented, ENERGETIC and hardworking, but respectful and kind-hearted!} 
         & \begin{tabular}{l} \\ \includegraphics[width=0.7\linewidth]{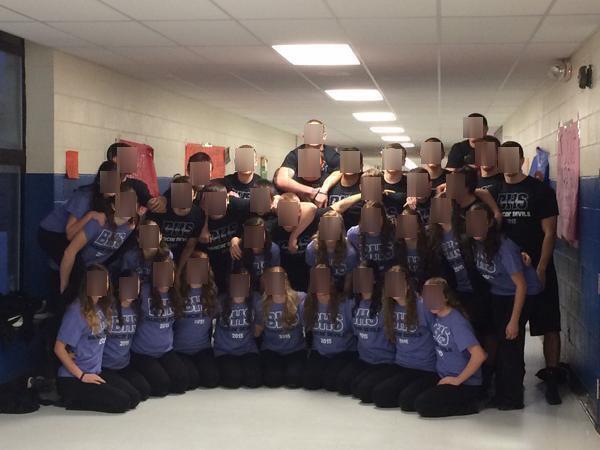} \\ \end{tabular}
         & positive 
         & \begin{tabular}{l}
             GPT-3:positive\\
             Flan-T5: positive\\
             IDEFICS: positive\\
             InstructBLIP: positive
         \end{tabular} 
         \\ \hline
         
         \textbf{MSD} & \scriptsize{Text: it's the insensitive strikeouts at suntrust park. \#braves \#chopchop} 
         & \begin{tabular}{l} \\ \includegraphics[width=0.7\linewidth]{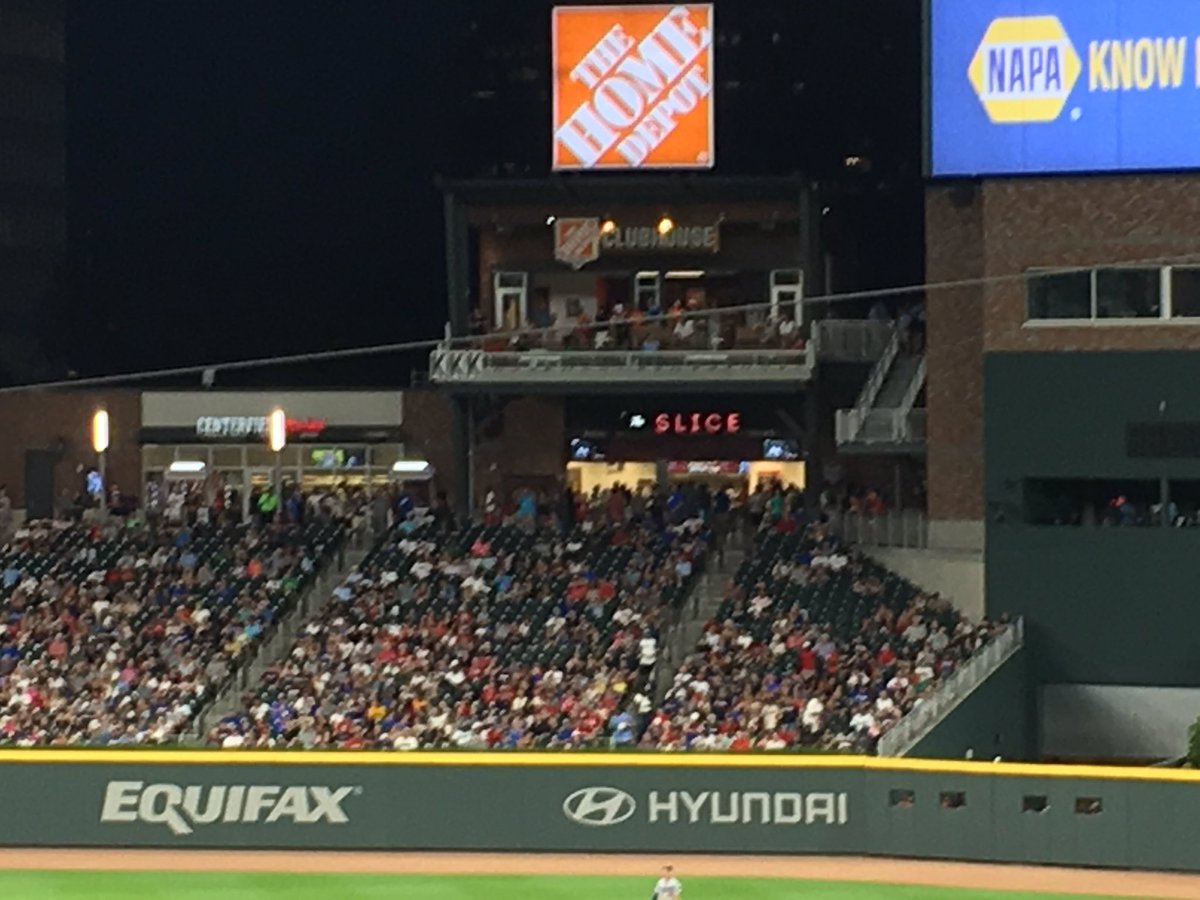} \\ \end{tabular}
         & sarcastic 
         & \begin{tabular}{l}
             GPT-3: sarcastic\\
             Flan-T5: sarcastic\\
             IDEFICS: not sarcastic\\
             InstructBLIP: not sarcastic
         \end{tabular} 
         \\ \hline         
          
    \end{tabularx}
    \caption{Text-Image examples and corresponding labels assigned by each LLM model for MVSA (sentiment analysis) and MSD (sarcasm detection) datasets. For each model we use the prompt templates included in Appendix \ref{appdx:prompting}.}
    \label{tab:ex_prompts}
\end{table*}
\section{Prompting}
\label{appdx:prompting}
For each dataset, we construct a prompt to include two randomly selected training examples for each class (GPT-3, FLAN-T5, IDEFICS) as follows:

\begin{itemize}
    \item TIR (GPT-3 \& FLAN-T5)

\begin{quote}
\small
\it
\noindent\textit{Label the next text as `image adds and text is represented', 'image adds and text is not represented', 'image does not add and text is represented', 'image does not add and text is not represented'. Text: <TWEET-TRAIN> // <LABEL-TRAIN>} \textbf{$\times 8$}

\noindent\textit{Label the next text as `image adds and text is represented', 'image adds and text is not represented', 'image does not add and text is represented', 'image does not add and text is not represented'. Text: <TWEET> //}

\end{quote}

\item TIR (IDEFICS)

\begin{quote}
\small
\it
\noindent\textit{User: <IMAGE-TRAIN> Label the image as `image adds and text is represented', `image adds and text is not represented', `image does not add and text is represented', `image does not add and text is not represented'. Assistant:<LABEL-TRAIN>} \textbf{$\times 8$} 

\noindent\textit{User: <IMAGE-TEST> Label the image as `image adds and text is represented', `image adds and text is not represented', `image does not add and text is represented', `image does not add and text is not represented'. Assistant:}

\end{quote}

\item TIR (InstructBLIP)

\begin{itemize}
\small{
    \item Prompt: \textit{Label the image as `image adds and text is represented', `image adds and text is not represented', `image does not add and text is represented', `image does not add and text is not represented'}
    \item Image: \textit{<IMAGE-TEST>}
    }
\end{itemize}

\item MVSA (GPT-3 \& FLAN-T5)

\begin{quote}
\small
\it
\noindent\textit{Label the next text as  `positive' or `negative' or `neutral'. Text: <TWEET-TRAIN> // <LABEL-TRAIN>} \textbf{$\times 6$}

\noindent\textit{Label the next text as `positive' or `negative' or `neutral'. Text: <TWEET> //}
\end{quote}

\item MVSA (IDEFICS)
\begin{quote}
\small
\it
\noindent\textit{User: <IMAGE-TRAIN> Is the sentiment of the image `positive' or `negative' or `neutral'?. Assistant:<LABEL-TRAIN>} \textbf{$\times 6$} 

\noindent\textit{User: <IMAGE-TEST> Is the sentiment of the image `positive' or `negative' or `neutral'?. Assistant:}
\end{quote}

\item MVSA (InstructBLIP)

\begin{itemize}
\small{
    \item Prompt: \textit{Is the sentiment of the image `positive' or `negative' or `neutral'?}
    \item Image: \textit{<IMAGE-TEST>}
    }
\end{itemize}

\item MHP

\begin{quote}
\small
\it
\noindent\textit{Label the next text as `hateful', `counterspeech', `reclaimed' or `none'. Text: <TWEET-TRAIN> // <LABEL-TRAIN>} \textbf{$\times 8$}

\noindent\textit{Label the next text as `hateful', `counterspeech', `reclaimed' or `none'. Text: <TWEET> //}
\end{quote}

\item MHP (IDEFICS)
\begin{quote}
\small
\it
\noindent\textit{User: <IMAGE-TRAIN> Is the image `hateful', `counterspeech', `reclaimed' or `none'?. Assistant:<LABEL-TRAIN>} \textbf{$\times 8$} 

\noindent\textit{User: <IMAGE-TEST> Is the image `hateful', `counterspeech', `reclaimed' or `none'?. Assistant:}
\end{quote}

\item MHP (InstructBLIP)

\begin{itemize}
\small{
    \item Prompt: \textit{Is the image `hateful', `counterspeech', `reclaimed' or `none'?}
    \item Image: \textit{<IMAGE-TEST>}
    }
\end{itemize}

\item MSD (GPT-3 \& FLAN-T5)

\begin{quote}
\small
\it
\noindent\textit{Label the next text as `sarcastic' or `not sarcastic'. Text: <TWEET-TRAIN> // <LABEL-TRAIN>} \textbf{$\times 4$}

\noindent\textit{Label the next text as  `sarcastic' or `not sarcastic'. Text: <TWEET> //}
\end{quote}

\item MSD (IDEFICS)
\begin{quote}
\small
\it
\noindent\textit{User: <IMAGE-TRAIN> Is the image `sarcastic' or `not sarcastic'? Assistant:<LABEL-TRAIN>} \textbf{$\times 4$} 

\noindent\textit{User: <IMAGE-TEST> Is the image `sarcastic' or `not sarcastic'? Assistant:}
\end{quote}


\begin{table*}[t!]
\tiny
\centering
\begin{tabular}{|l|cc|cc|cc|cc|cc|}
\hline
\textbf{Dataset}        & \multicolumn{2}{c|}{\textbf{TIR}}                                & \multicolumn{2}{c|}{\textbf{MVSA}}                & \multicolumn{2}{c|}{\textbf{MHP}}         & \multicolumn{2}{c|}{\textbf{MSD}} & \multicolumn{2}{c|}{\textbf{MICD}}       \\ \hline
            & $\lambda_1$, $\lambda_2$, $\lambda_3$ & E                         & $\lambda_1$, $\lambda_2$, $\lambda_3$ & E          & $\lambda_1$, $\lambda_2$, $\lambda_3$ & E  & $\lambda_1$, $\lambda_2$, $\lambda_3$ & E & $\lambda_1$, $\lambda_2$, $\lambda_3$ & E \\ \hline
{Ber-ViT-Conc}   & -                                   & 3                         & -                                   & 7          & -                                   & 7  & -                                   & 1 & - & 2 \\
{Ber-ViT-Conc+C}                     & 0.9, 0.1, 0                           & 3                         & 0.9, 0.1, 0                           & 5          & 0.9, 0.1, 0                           & 7  & 0.9, 0.1,0                           & 6 & 0.9,0.1,0 & 2 \\
{Ber-ViT-Conc+M}                     & 0.9, 0, 0.1                           & 4                         & 0.9, 0, 0.1                           & 6          & 0.9, 0, 0.1                           & 9  & 0.9, 0, 0.1                           & 3 & 0.9,0,0.1 & 1 \\
{Ber-ViT-Conc+C+M}                       & 0.8, 0.1, 0.1                         & 6                         & 0.8, 0.1, 0.1                         & 4          & 0.8, 0.1, 0.1                         & 6  & 0.8, 0.1, 0.1                         & 3 &  0.8,0.1,0.1 & 2 \\
{Ber-ViT-Att}    & -                                   & 2                         & -                                   & 8          & -                                   & 7  & -                                   & 1 & - & 3 \\
{Ber-ViT-Att+C}                     & 0.9, 0.1,0                           & 2                         & 0.9, 0.1, 0                           & 8          & 0.9,0.1,0                           & 7  & 0.9, 0.1, 0                           & 3 & 0.9,0.1,0 & 2 \\
{Ber-ViT-Att+M}                     & 0.92, 0, 0.08                       & 3                         & 0.9, 0, 0.1                           & 6          & 0.9,0,0.1                           & 6  & 0.9, 0, 0.1                           & 3 & 0.9,0,0.1 & 1 \\
{Ber-ViT-Att+C+M}                      & 0.8, 0.1, 0.1                         & 4                         & 0.8, 0.1, 0.1                         & 15         & 0.8,0.1,0.1                         & 13 & 0.8, 0.1, 0.1                         & 5 & 0.8,0.1,0.1 & 2 \\
{MMBT}           & -                                   & 2                         & -                                   & 9          & -                                   & 5  & -                                   & 1 & - & 1 \\
{MMBT+C}                     & 0.9, 0.1, 0                           & 4                         & 0.9, 0.1, 0                           & {5} & 0.9, 0.1, 0                           & 9  & 0.9,0.1,0                           & 3 &0.9,0.1,0 & 2 \\
{MMBT+M}                     & 0.9, 0, 0.1                           & 4                         & 0.7, 0 ,0.3                        & 6          & 0.9, 0, 0.1                           & 9  & 0.82, 0, 0.08                       & 4 & 0.9,0,0.1 & 2 \\
{MMBT+C+M}                      & 0.84, 0.08, 0.08                      & 3                         & 0.85, 0.1, 0.05                       & 11         & 0.8, 0.1, 0.1                         & 10 & 0.85,0.1,0.05                       & 3 & 0.6,0.2,0.2 & 4\\
{LXMERT}         & -                                   & 2                         & -                                   & 5          & -                                   & 5  & {-}                          & 2 & - & 3 \\
{LXMERT+C}                      & 0.9,0.1,0                           & 2                         & 0.9,0.1,0                           & 8          & 0.9, 0.1, 0                           & 5  & 0.9,0.1,0                           & 2 & 0.9,0.1,0 & 2 \\
{LXMERT+M}                     & 0.85,0,0.15                       & 1                         & 0.9,0,0.1                           & 6          & 0.8, 0, 0.1                           & 12 & 0.85,0,0.15                       & 2 & 0.9,0,0.1 & 3 \\
{LXMERT+C+M}                      & 0.9, 0.08, 0.02                       & 2                         & 0.83,0.02,0.15                      & {7} & 0.8, 0.1, 0.1                         & 11 & 0.85, 0.1, 0.05                       & 2 & 0.8,0.1,0.1 & 3 \\
{ViLT}           & -                                   & 6                         & -                                   & 5          & -                                   & 4  & -                                   & 1 & - & 4 \\
{ViLT+C}                      & 0.9, 0.1, 0                           & 6                         & 0.9, 0.1, 0                           & 11         & 0.9, 0.1, 0                           & 4  & 0.9, 0.1, 0                           & 1 & 0.95,0.05,0 & 2 \\
{ViLT+M}                      & 0.85, 0, 0.15                       & 5                         & 0.9,0,0.1                           & 3          & 0.9, 0, 0.1                           & 7  & 0.9, 0, 0.1                           & 2 & 0.92,0,0.08 & 2 \\
{ViLT+C+M}                       & 0.8, 0.1, 0.1                         & 2                         & 0.8, 0.1, 0.1                         & 13         & 0.8, 0.1, 0.1                        & 9  & 0.8, 0.1, 0.1                         & 2 & 0.87,0.05,0.08 & 1 \\ \hline 

\end{tabular}
\caption{Hyperaprameter values for $\lambda_1$, $\lambda_2$, $\lambda_3$ as explained in Section \ref{sec:aux-tasks}, and number of fine-tuning epochs (E) for each model.}
\label{tab:hp-lambda}
\end{table*}

\item MSD (InstructBLIP)

\begin{itemize}
\small{
    \item Prompt: \textit{Is the image `sarcastic' or `not sarcastic'?}
    \item Image: \textit{<IMAGE-TEST>}
    }
\end{itemize}

\item MICD (GPT-3 \& FLAN-T5)
\begin{quote}
\small
\it
\noindent\textit{Label the next text as `commercial' or `not commercial'. Text: <TWEET-TRAIN> // <LABEL-TRAIN>} \textbf{$\times 4$}

\noindent\textit{Label the next text as `commercial' or `not commercial'. Text: <TWEET> //}
\end{quote}

\item MICD (IDEFICS)
\begin{quote}
\small
\it
\noindent\textit{User: <IMAGE-TRAIN> Is the image `commercial' or `not commercial'? Assistant:<LABEL-TRAIN>} \textbf{$\times 4$} 

\noindent\textit{User: <IMAGE-TEST> Is the image `commercial' or `not commercial'? Assistant:}
\end{quote}

\item MICD (InstructBLIP)
\small
\begin{itemize}
    \item Prompt: \textit{Is the image `commercial' or `not commercial'?}
    \item Image: \textit{<IMAGE-TEST>}
\end{itemize}

\end{itemize}
\noindent\small <Label-TRAIN> \normalsize corresponds to the true label of the \small<TWEET-TRAIN> \normalsize 
training example, \small<TWEET> \normalsize refers to a testing example. We remove punctuation and spaces and map the output of each model (FLAN-T5 or GPT-3) to the corresponding label. Table \ref{tab:ex_prompts} shows examples of outputs for each LLM model for MVSA and MSD datasets.

\subsection{Implementation Details}
\paragraph{FLAN-T5 \& IDEFICS} We use one GPU T4 to obtain the inference results from Flan-T5 \citep{chung2022scaling} and IDEFICS \citep{laurenccon2023obelisc} models. For Flan-T5 we use the large version from the Hugging Face library (780M parameters) \citep{wolf2019huggingface}. For IDEFICS, we use the 9B parameters instruct version of the model (\textit{idefics-9b-instruct}) via Hugging Face library.
\paragraph{InstructBLIP} We use one A100 GPU to obtain inference results from InstructBLIP \cite{dai2023instructblip}. We use the 7B-parameters version (\textit{instructblip-vicuna-7b}) from the Hugging Face library.
\paragraph{GPT-3} For GPT-3 \cite{brown2020language}, we use the \textit{text-davinci-003} model via the OpenAI\footnote{\url{https://platform.openai.com/docs/api-reference}} Library.
\paragraph{Note on GPT-4} For this work, we opted not to include GPT-4 due to (1) its nature as a black-box model accessible only through a paid API; (2) the lack of information regarding the pre-training data, raising concerns about potential exposure to the test sets and thus, information leakage.

\end{document}